\documentclass{article}
\usepackage{spconf,amsmath,graphicx,hyperref}
\usepackage{booktabs}
\usepackage{longtable}
\usepackage{amsmath}
\usepackage{amssymb}
\usepackage{graphicx}
\usepackage{xcolor}
\usepackage{lineno}
\usepackage{float}
\usepackage[ruled,vlined,linesnumbered,algo2e]{algorithm2e}
\usepackage[OT1]{fontenc}
\usepackage{microtype}
\hypersetup{hidelinks}

\newcommand{\method}{CUBE}
\newcommand{\methode}{CUBE-E}
\newcommand{\methodfull}{Class-conditioned Unknown Boundary Evidence}

\title{CUBE: Class-conditioned Unknown Boundary Evidence for Open-Set Insect Species Recognition}
\name{
Hao Tan$^{1,2,*}$,
Qimin Zhang$^{3,*}$,
Zhe Xiao$^{4}$,
Zekai Zuo$^{1}$,
Wanting Liu$^{1}$,
Jinghui Qin$^{1,\dagger}$}

\address{
$^{1}$Guangdong University of Technology \\
$^{2}$The Hong Kong University of Science and Technology (Guangzhou)\\
$^{3}$The Hong Kong Polytechnic University
\quad
$^{4}$Zhuhai Campus, Zunyi Medical University\\
}
\begin{document}
\maketitle
\begingroup
\renewcommand{\thefootnote}{\fnsymbol{footnote}}
\footnotetext[1]{These authors contributed equally.}
\footnotetext[2]{Corresponding author.}
\endgroup
\begin{abstract}

Open-set insect species recognition requires models capable of detecting and categorizing unknown or novel species that are never seen during training but often resemble known classes.
Unlike generic out-of-distribution samples, the feature representations of these unknown species may lie near a particular known-species boundary and induce confident errors. 
To address this, we propose \textbf{\methodfull{} (\method{})}, which models unknown species locally around each known species rather than as a single global unknown class. 
\method{} exploits auxiliary species to learn class-conditioned boundaries and preserves their species prototypes as shadow references during inference. It further combines visual boundary evidence with taxonomic evidence and candidate calibration.
When location metadata are available, we propose \methode{} to incorporate ecological consistency.
Extensive experiments on Open-Insect C-America show that our method achieves state-of-the-art performance, reaching 98.4\% mean AUROC and 95.2\% closed-set accuracy.

\end{abstract}
\section{Introduction}
\label{sec:intro}

Automated species recognition has become an important tool for biodiversity monitoring, ecological surveys, and large-scale insect observation~\cite{jain2024insect,gharaee2024bioscan,chen2026open}. However, biodiversity data are inherently open-ended, and deployed systems inevitably encounter species absent from the training set. A robust recognition system should recognize known species while reliably rejecting unknown ones. This is particularly difficult for insects because unknown species are often not visually anomalous, but local unknowns may share wing shapes, textures, and colors with known species. Thus, Open-set insect recognition is a fine-grained boundary problem rather than simple anomaly detection~\cite{lang2024coarse,chen2026open}.

Existing out-of-distribution (OOD) and open-set recognition methods mainly rely on post-hoc confidence or feature scores, training regularization, or auxiliary outliers~\cite{liu2023fast,Ling2025CADRefRO,liu2025detecting,wei2022logitnorm}. Despite their effectiveness, many methods still model unknown species globally: an unknown sample should have low confidence, lie far from all known classes, or belong to a shared negative group. This assumption is poorly matched to fine-grained species recognition. The hardest unknown insects may be close to only a few known species and can be confidently absorbed by one of them. Rejecting such samples requires estimating the credible boundary of the particular known species they resemble, rather than measuring deviation from the known distribution as a whole.
Moreover, auxiliary species provide a natural means of learning these local boundaries, but treating them as an undifferentiated negative pool discards useful information. Samples from the same auxiliary species form coherent visual groups, while each auxiliary species is typically confusable with only a small subset of known species. Therefore, they can serve as boundary probes during training and as structured nearby-unknown references during inference~\cite{zhang2023mixture,li2024all,regmi2026image}. Besides, species recognition offers biological context unavailable in generic OOD detection. Taxonomic relations provide semantic cues about biological similarity~\cite{stevens2024bioclip,yang2024biotrove,gu2026bioclip}, while location metadata reflects species distributions~\cite{gharaee2024bioscan,chen2026open}. These cues can help verify whether a visually plausible prediction is also biologically plausible.

\begin{figure*}[t]
\centering
\includegraphics[width=0.8\textwidth]{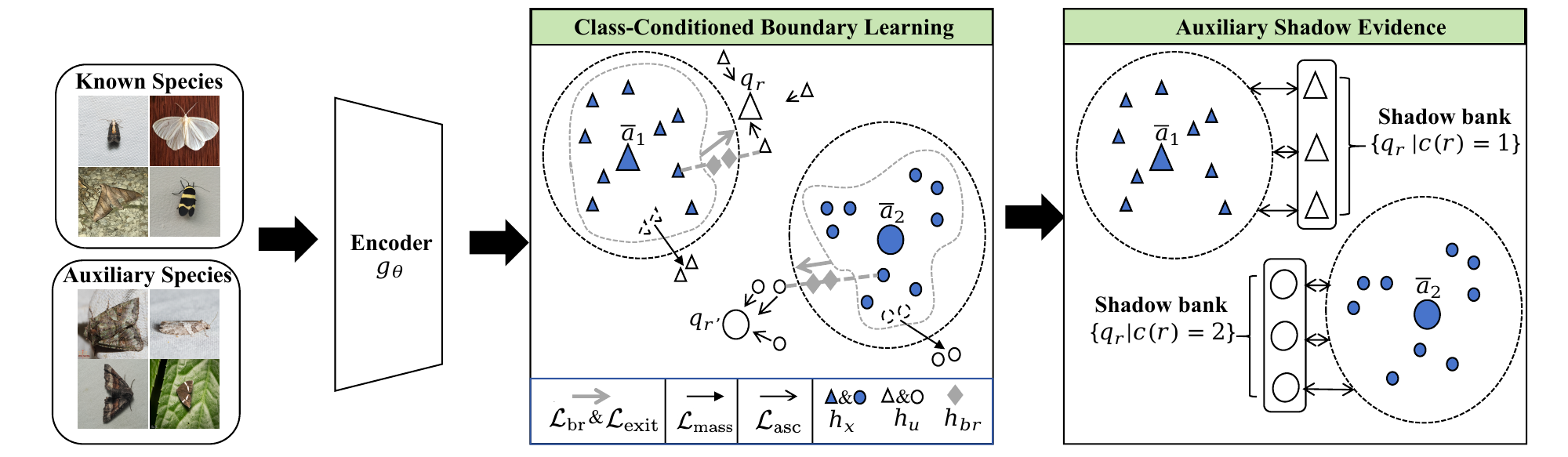}
\caption{Class-conditioned boundary learning and auxiliary shadow-bank construction in CUBE.}
\label{fig:arh}
\end{figure*}

Therefore, we propose \textbf{\methodfull{}} (\textbf{\method{}}), a class-conditioned framework that judges each fine-grained unknown species relative to the known species it most closely resembles. \method{} associates auxiliary samples with locally confusable known species to learn rejection boundaries and preserves auxiliary prototypes as inference-time shadow references. 
Taxonomic candidate calibration is introduced to further reduce confusion among plausible known species, while \methode{} optionally incorporates ecological consistency.
Extensive experiments on Open-Insect C-America demonstrate state-of-the-art (SOTA) OOD detection performance while preserving strong closed-set recognition performance.
Overall, our contributions are two-fold. First, we formulate fine-grained unknown detection as class-conditioned boundary estimation instead of global unknown species modeling. Second, we develop structured auxiliary boundary evidence that unifies local boundary learning and inference-time shadow referencing, leading to SOTA open-set recognition performance.

\section{Proposed Method}
\label{sec:method}

\subsection{Problem Formulation and Overview}

We consider open-set fine-grained insect species recognition with an In-Distribution (ID) training set $\mathcal{D}_{\mathrm{id}}=\{(x_i,y_i)\}_{i=1}^{N}$, where $y_i\in\{1,\ldots, C\}$ denotes one of $C$ known species, and an auxiliary set $\mathcal{D}_{\mathrm{aux}}=\{(u_j,r_j)\}_{j=1}^{N_{\mathrm{aux}}}$, where $r_j$ is the auxiliary species identity of $u_j$. Auxiliary species are not added to the closed-set label space. Instead, they serve as structured probes of visually similar non-ID species around known classes. An encoder $g_\theta$ maps an image to the normalized feature $h=\operatorname{Norm}(g_\theta(x))$. A $C$-way linear classifier produces $\mathbf{z}(h)=[z_1(h),\ldots,z_C(h)]^\top$, where $z_c(h)=w_c^\top h+b_c$ is the closed-set logit, with $p(c|x)=[\operatorname{softmax}(\mathbf{z}(h))]_c$ and $\hat y=\arg\max_c p(c|x)$. CUBE follows a class-conditioned principle: a hard unknown species need not be far from all known species, but should lie outside the credible boundary of the particular species it resembles, so CUBE learns class-conditioned boundary evidence, preserves auxiliary prototypes as shadow references, and incorporates taxonomic evidence and candidate calibration. If location metadata is available, CUBE can optionally add ecological context. We name this variant CUBE-E.

\subsection{Class-Conditioned Boundary Evidence}

As illustrated in Fig.~\ref{fig:arh}, for each known species $c$, CUBE maintains a support bank $\mathcal{A}_c=\{a_{c}^1,\ldots, a_{c}^S\}$ of $S$ normalized anchors, updated by exponential moving averages of training features $h_x$, with class mean $\bar a_c=S^{-1}\sum_{s=1}^{S}a_{c}^s$. Given an auxiliary image $u$, CUBE identifies the known species most likely to absorb it:
\begin{equation}
\small
s_c(u)=\max_{a\in\mathcal{A}_c}
\operatorname{Norm}(g_\theta(u))^\top a,
\quad
\mathcal{N}_K(u)=\operatorname{TopK}_{c}\,s_c(u).
\end{equation}
Here, $\mathcal{N}_K(u)$ contains the $K$ known species with the largest support similarities. CUBE suppresses the closed-set probability mass assigned to this local neighborhood via:
\begin{equation}
\small
\scalebox{0.96}{$\displaystyle
\mathcal{L}_{\mathrm{mass}}
=
\mathbb{E}_{u\in\mathcal{D}_{\mathrm{aux}}}
\left[
\omega(u)
\left(
-\log\left(
1-\sum_{c\in\mathcal{N}_K(u)}p(c|u)
\right)
\right)
\right],
$}
\end{equation}
where the non-negative $\omega(u)$ increases with $\max_c s_c(u)$, emphasizing auxiliary samples near known-species boundaries.  

To provide an explicit local rejection direction, each known species has an exit logit $e_c(h)=(w_c^{\mathrm{ex}})^\top h+b_c^{\mathrm{ex}}$, parallel to its closed-set logit. Then, the local unknown probability $\pi_c(h)$ can be computed as $\exp(e_c(h))/(\exp(z_c(h))+\exp(e_c(h)))$. Known samples deactivate the exit of their ground-truth class, whereas auxiliary samples activate the exits of classes in $\mathcal{N}_K(u)$. This is achieved by the following learning objective:
\begin{equation}
\small
\begin{aligned}
\mathcal{L}_{\mathrm{exit}}
={}&
\mathbb{E}_{(x,y)\sim\mathcal{D}_{\mathrm{id}}}
\left[\ell_{\mathrm{BCE}}\bigl(\pi_y(h_x),0\bigr)\right]\\
&+
\mathbb{E}_{u\sim\mathcal{D}_{\mathrm{aux}}}
\left[
\frac{1}{|\mathcal{N}_K(u)|}
\sum_{c\in\mathcal{N}_K(u)}
\ell_{\mathrm{BCE}}\bigl(\pi_c(h_u),1\bigr)
\right],
\end{aligned}
\end{equation}
where $\ell_{\mathrm{BCE}}(\pi,t)=-t\log\pi-(1-t)\log(1-\pi)$. To smooth the transition between known and nearby unknown regions, CUBE constructs $h_{\mathrm{br}}=\operatorname{Norm}(\rho h_x+(1-\rho)h_u)$ with a soft unknown target $1-\rho$ and optimizes the following objective:
\begin{equation}
\small
\mathcal{L}_{\mathrm{br}}=\mathbb{E}_{(x,y),u,\rho}[\ell_{\mathrm{BCE}}(\pi_y(h_{\mathrm{br}}),1-\rho)].
\end{equation}
CUBE further preserves the species-level structure of auxiliary data. Let $\mathcal{D}_r$ contain auxiliary images of species $r$, and define its prototype as $q_r=\operatorname{Norm}(|\mathcal{D}_r|^{-1}\sum_{u\in\mathcal{D}_r}g_\theta(u))$. Auxiliary species consistency loss attracts each $h_u$ to its corresponding $q_r$ while separating it from other auxiliary prototypes and known supports:
\begin{equation}
\small
\mathcal{L}_{\mathrm{asc}}
=
-\mathbb{E}_{(u,r)}
\log
\frac{\exp(h_u^\top q_r/\tau)}
{\sum_{r'}\exp(h_u^\top q_{r'}/\tau)
+\sum_c\exp(h_u^\top\bar a_c/\tau)},
\end{equation}
where $h_u=\operatorname{Norm}(g_\theta(u))$ and $\tau$ is the temperature. The overall training objective is
\begin{equation}
\small
\mathcal{L}
=
\mathcal{L}_{\mathrm{id}}
+\lambda_{\mathrm{bd}}
\left(
\mathcal{L}_{\mathrm{mass}}
+\lambda_{\mathrm{exit}}\mathcal{L}_{\mathrm{exit}}
+\lambda_{\mathrm{br}}\mathcal{L}_{\mathrm{br}}
+\lambda_{\mathrm{asc}}\mathcal{L}_{\mathrm{asc}}
\right),
\end{equation}
where $\mathcal{L}_{\mathrm{id}}=-\mathbb{E}_{(x,y)\in\mathcal{D}_{\mathrm{id}}}\log p(y|x)$, and the $\lambda$ terms are non-negative balancing coefficients.

\subsection{Auxiliary Shadow Evidence}

After training, CUBE reuses the auxiliary prototypes as an inference-time shadow bank, as shown in Fig.~\ref{fig:arh}. Each prototype $q_r$ is attached to its nearest known species by $c(r)=\arg\max_c q_r^\top\bar a_c$. For a test image $x$ with visual prediction $\hat y$, CUBE constructs a candidate set $\mathcal{C}(x)$ from the top visual and support-neighbor species, and considers the shadow prototypes attached to these candidates:
\begin{equation}
\small
E_{\mathrm{sh}}(x)
=
\left[
\max_{r:\,c(r)\in\mathcal{C}(x)}
h^\top q_r
-h^\top\bar a_{\hat y}
+\delta_{\mathrm{sh}}
\right]_+ .
\end{equation}
Here, $[\cdot]_+=\max(\cdot,0)$, and $\delta_{\mathrm{sh}}$ is a margin. Positive shadow evidence indicates that $x$ is closer to a nearby auxiliary unknown reference than to the mean support of its predicted known species.

\subsection{Taxonomic Candidate Calibration}

CUBE uses a frozen biological vision-language model to evaluate whether a visual prediction is taxonomically plausible. For each known species $c$, its species, genus, and family names are encoded into a text prototype $t_c$. The frozen image encoder produces $v(x)$, and their agreement is $b_c(x)=\operatorname{Norm}(v(x))^\top\operatorname{Norm}(t_c)$. Taxonomic evidence for rejection is
\begin{equation}
\small
E_{\mathrm{tax}}(x)
=
\left[\max_c b_c(x)-b_{\hat y}(x)\right]_+
+\left[\tau_{\mathrm{tax}}-b_{\hat y}(x)\right]_+ ,
\end{equation}
where $b_{\hat y}(x)$ is the agreement of the visually predicted species and $\tau_{\mathrm{tax}}$ is the minimum semantic-support threshold. These two terms capture disagreement with a more plausible species and insufficient absolute support, respectively.

Taxonomy also calibrates closed-set recognition within the visually plausible candidates. Let $\operatorname{TopM}(\mathbf{z}(h))$ denote the $M$ species with the largest visual logits. For each $c$ in this set, CUBE computes
\begin{equation}
\small
\tilde z_c(x)
=
z_c(h)
+\eta_{\mathrm{sup}}\Delta_{\mathrm{sup}}(x,c)
+\eta_{\mathrm{tax}}b_c(x).
\end{equation}
Here, $\Delta_{\mathrm{sup}}(x,c)=R^{-1}\sum_{a\in\mathcal{T}_R(x,c)}h^\top a$, where $\mathcal{T}_R(x,c)$ contains the $R$ anchors in $\mathcal{A}_c$ most similar to $h$. The coefficients $\eta_{\mathrm{sup}}$ and $\eta_{\mathrm{tax}}$ control support and taxonomic calibration, and the final closed-set prediction is selected from these calibrated top-$M$ logits.

\subsection{Ecological Context and Inference}

When location metadata are available, we further propose CUBE-E, an extension of CUBE that evaluates the geographical plausibility of a prediction without modifying the visual encoder. Given the latitude-longitude vector $\ell(x)$, let $m_c,\sigma_c$ denote the geographical mean and scale of species $c$, and let $m_0,\sigma_0$ denote the corresponding global statistics. The geographical mismatch is defined as follows:
\begin{equation}
\small
d_{\mathrm{geo}}(x,c)
=
\beta
\left\|
\frac{\ell(x)-m_c}{\sigma_c}
\right\|_2
+
(1-\beta)
\left\|
\frac{\ell(x)-m_0}{\sigma_0}
\right\|_2 ,
\end{equation}
where the divisions are element-wise and $\beta\in[0,1]$ balances class-specific and global geography. Classes with insufficient location samples use only the global term. For $c\in\operatorname{TopM}(\mathbf{z}(h))$, CUBE-E computes $\tilde z_c^{\mathrm{ctx}}(x)=\tilde z_c(x)-\eta_{\mathrm{geo}}d_{\mathrm{geo}}(x,c)$, where $\eta_{\mathrm{geo}}$ controls geographical calibration. Its ecological OOD evidence  $E_{\mathrm{geo}}(x)$ is obtained by computing $d_{\mathrm{geo}}(x,\hat y)$.

\begin{table}[!t]
\centering
\caption{Main results on Open-Insect C-America. $\dagger$ indicates a statistically significant improvement over the strongest baseline for each metric under a two-sided paired $t$-test ($p<0.05$).}
\label{tab:camerica_main}
\small
\resizebox{\columnwidth}{!}{%
\begin{tabular}{lcccc}
\toprule
Method  & L $\uparrow$ & NL $\uparrow$ & NM $\uparrow$ & Acc. \\
\midrule
ASCOOD (CVPR '26) & 83.3{\scriptsize$\pm0.9$} & 83.5{\scriptsize$\pm1.1$} & 85.5{\scriptsize$\pm0.8$} & 91.5{\scriptsize$\pm0.4$} \\
CADRef  (CVPR '25) & 90.6{\scriptsize$\pm0.4$} & 92.4{\scriptsize$\pm0.6$} & 96.4{\scriptsize$\pm1.3$} & 93.7{\scriptsize$\pm0.2$} \\
Mahalanobis++  (ICML '25) & 88.0{\scriptsize$\pm0.8$} & 88.7{\scriptsize$\pm1.0$} & 97.4{\scriptsize$\pm1.7$} & 93.7{\scriptsize$\pm0.2$} \\
OODD  (CVPR '25) & 82.8{\scriptsize$\pm0.7$} & 85.3{\scriptsize$\pm0.9$} & 96.5{\scriptsize$\pm0.5$} & 93.7{\scriptsize$\pm0.4$} \\
NECO  (ICLR '24) & 79.5{\scriptsize$\pm1.1$} & 78.9{\scriptsize$\pm1.4$} & 95.5{\scriptsize$\pm0.6$} & 93.7{\scriptsize$\pm0.4$} \\
NCI  (CVPR '25) & 92.4{\scriptsize$\pm0.6$} & 94.1{\scriptsize$\pm0.7$} & 96.9{\scriptsize$\pm0.4$} & 93.7{\scriptsize$\pm0.2$} \\
FDBD  (ICML '24) & 92.7{\scriptsize$\pm0.5$} & 94.3{\scriptsize$\pm0.7$} & 96.2{\scriptsize$\pm0.2$} & 93.7{\scriptsize$\pm0.5$} \\
NovelBranch  (NeurIPS '25) & 94.0{\scriptsize$\pm0.5$} & 97.9{\scriptsize$\pm1.2$} & 97.3{\scriptsize$\pm0.3$} & 94.2{\scriptsize$\pm0.3$} \\
MixOE  (WACV '23) & 95.0{\scriptsize$\pm0.4$} & 98.1{\scriptsize$\pm0.3$} & 98.1{\scriptsize$\pm1.1$} & 94.5{\scriptsize$\pm0.4$} \\
Extended  (NeurIPS '25) & 95.0{\scriptsize$\pm0.4$} & 98.3{\scriptsize$\pm0.3$} & 98.3{\scriptsize$\pm0.4$} & 94.4{\scriptsize$\pm0.7$} \\
\midrule
\method{}  (Ours) & \underline{95.8{\scriptsize$\pm0.4$}}\textsuperscript{$\dagger$} & \underline{98.7{\scriptsize$\pm0.1$}}\textsuperscript{$\dagger$} & \underline{99.1{\scriptsize$\pm0.2$}}\textsuperscript{$\dagger$} & \underline{94.9{\scriptsize$\pm0.2$}}\textsuperscript{$\dagger$} \\
\methode{}  (Ours) & \textbf{96.0{\scriptsize$\pm0.3$}}\textsuperscript{$\dagger$} & \textbf{99.7{\scriptsize$\pm0.2$}}\textsuperscript{$\dagger$} & \textbf{99.5{\scriptsize$\pm0.1$}}\textsuperscript{$\dagger$} & \textbf{95.2{\scriptsize$\pm0.3$}}\textsuperscript{$\dagger$} \\
\bottomrule
\end{tabular}
}
\end{table}

\begin{table}[!t]
\centering
\caption{Cross-source evaluation on the BCI OOD split. $\dagger$ indicates a statistically significant improvement over the strongest baseline for each metric under a two-sided paired $t$-test ($p<0.05$).}
\label{tab:bci}
\resizebox{\columnwidth}{!}{%
\begin{tabular}{lcccc}
\toprule
Method  & AUROC $\uparrow$ & AUPR $\uparrow$& FPR95 $\downarrow$& Acc. \\
\midrule
ASCOOD (CVPR '26) & 81.1{\scriptsize$\pm$1.4} & 11.1{\scriptsize$\pm$1.1} & 81.5{\scriptsize$\pm$3.6} & 91.5{\scriptsize$\pm$0.4} \\
CADRef (CVPR '25) & 93.1{\scriptsize$\pm$0.7} & 10.2{\scriptsize$\pm$0.9} & 25.1{\scriptsize$\pm$2.1} & 93.7{\scriptsize$\pm$0.2} \\
Mahalanobis++ (ICML '25) & 90.9{\scriptsize$\pm$1.2} & 5.9{\scriptsize$\pm$0.8} & 22.0{\scriptsize$\pm$2.8} & 93.7{\scriptsize$\pm$0.2} \\
OODD (CVPR '25) & 78.0{\scriptsize$\pm$1.8} & 2.2{\scriptsize$\pm$0.6} & 40.5{\scriptsize$\pm$4.2} & 93.7{\scriptsize$\pm$0.4} \\
NECO (ICLR '24) & 84.8{\scriptsize$\pm$0.6} & 3.6{\scriptsize$\pm$0.9} & 36.7{\scriptsize$\pm$3.4} & 93.7{\scriptsize$\pm$0.4} \\
NCI (CVPR '25) & 93.9{\scriptsize$\pm$0.8} & 16.8{\scriptsize$\pm$1.5} & 26.2{\scriptsize$\pm$2.2} & 93.7{\scriptsize$\pm$0.2} \\
FDBD (ICML '24) & 94.4{\scriptsize$\pm$0.6} & 19.0{\scriptsize$\pm$1.8} & 25.2{\scriptsize$\pm$2.6} & 93.7{\scriptsize$\pm$0.5} \\
NovelBranch (NeurIPS '25) & 94.3{\scriptsize$\pm$1.2} & 24.5{\scriptsize$\pm$2.1} & 25.2{\scriptsize$\pm$1.9} & 94.2{\scriptsize$\pm$0.3} \\
MixOE (WACV '23) & 95.0{\scriptsize$\pm$0.4} & 32.1{\scriptsize$\pm$1.4} & \underline{17.9{\scriptsize$\pm$1.2}} & 94.5{\scriptsize$\pm$0.4} \\
Extended (NeurIPS '25) & 95.5{\scriptsize$\pm$0.3} & 34.4{\scriptsize$\pm$1.1} & 22.8{\scriptsize$\pm$1.7} & 94.4{\scriptsize$\pm$0.7} \\
\midrule
\method{} (Ours) & \textbf{96.1{\scriptsize$\pm$0.3}}\textsuperscript{$\dagger$} & \textbf{38.8{\scriptsize$\pm$0.8}}\textsuperscript{$\dagger$} & \textbf{16.1{\scriptsize$\pm$1.2}}\textsuperscript{$\dagger$} & \underline{94.9{\scriptsize$\pm$0.2}}\textsuperscript{$\dagger$} \\
\methode{} (Ours) & \underline{95.7{\scriptsize$\pm$0.2}}\textsuperscript{$\dagger$} & \underline{36.0{\scriptsize$\pm$1.3}}\textsuperscript{$\dagger$} & 19.5{\scriptsize$\pm$1.1} & \textbf{95.2{\scriptsize$\pm$0.3}}\textsuperscript{$\dagger$} \\
\bottomrule
\end{tabular}
}
\end{table}

\begin{table*}[t]
\centering
\caption{Module ablation on C-America. All variants use ConvNeXt-T. Aux., Sup., and Cons. denote coarse auxiliary boundary, support-split boundary, and auxiliary species consistency. Calib. denotes taxonomic candidate calibration.}
\label{tab:module_ablation}
\small
\begin{tabular}{lccccccccccc}
\toprule
Setting & \multicolumn{3}{c}{Boundary} & Shadow & \multicolumn{2}{c}{Taxonomy} & Eco & \multicolumn{4}{c}{Metric} \\
\cmidrule(lr){2-4}\cmidrule(lr){6-7}\cmidrule(lr){9-12}
 & Aux. & Sup. & Cons. & & Score & Calib. & & L & NL & NM & Acc. \\
\midrule
ID-only & -- & -- & -- & -- & -- & -- & -- & 92.9 & 94.7 & 96.7 & 93.7 \\
+ Coarse auxiliary boundary & $\checkmark$ & -- & -- & -- & -- & -- & -- & 94.9 & 97.9 & 98.1 & 93.5 \\
+ Support-split boundary & $\checkmark$ & $\checkmark$ & -- & -- & -- & -- & -- & 95.6 & 98.5 & 98.4 & 94.1 \\
+ Auxiliary species consistency & $\checkmark$ & $\checkmark$ & $\checkmark$ & -- & -- & -- & -- & 95.7 & 98.6 & 98.4 & 94.4 \\
+ Auxiliary shadow evidence & $\checkmark$ & $\checkmark$ & $\checkmark$ & $\checkmark$ & -- & -- & -- & 95.9 & 98.7 & 98.5 & 94.5 \\
+ Taxonomic scoring & $\checkmark$ & $\checkmark$ & $\checkmark$ & $\checkmark$ & $\checkmark$ & -- & -- & 95.8 & 98.7 & 99.2 & 94.5 \\
+ Taxonomic candidate calibration & $\checkmark$ & $\checkmark$ & $\checkmark$ & $\checkmark$ & $\checkmark$ & $\checkmark$ & -- & 95.8 & 98.7 & 99.2 & 94.7 \\
\method{} & $\checkmark$ & $\checkmark$ & $\checkmark$ & $\checkmark$ & $\checkmark$ & $\checkmark$ & -- & \underline{95.8} & \underline{98.7} & 99.1 & \underline{94.9} \\
\methode{} & $\checkmark$ & $\checkmark$ & $\checkmark$ & $\checkmark$ & $\checkmark$ & $\checkmark$ & $\checkmark$ & \textbf{96.0} & \textbf{99.7} & \textbf{99.5} & \textbf{95.2} \\
\bottomrule
\end{tabular}%
\end{table*}

For inference, the base visual score is $S_{\mathrm{base}}(x)=-\log p(\hat y|x)$ ${}+\gamma(1-h^\top\bar a_{\hat y})$, where $\gamma$ weights known-species support consistency. The local exit evidence is $E_{\mathrm{exit}}(x)=\sum_{c\in\mathcal{C}(x)}\kappa_c(x)\pi_c(h)$, where $\kappa_c(x)$ is a similarity-based weight normalized over $\mathcal{C}(x)$. To align the scales of different evidence sources, CUBE uses $\operatorname{Std}_{\mathrm{id}}(F(x))=(F(x)-\mu_F)/(\sigma_F+\epsilon)$, where $\mu_F$ and $\sigma_F$ are estimated from validation ID samples and $\epsilon$ is a numerical constant.
The final score is $S_{\mathrm{CUBE}}(x)=\operatorname{Std}_{\mathrm{id}}(S_{\mathrm{base}}(x))$ ${}+\alpha_{\mathrm{exit}}\operatorname{Std}_{\mathrm{id}}(E_{\mathrm{exit}}(x))$ ${}+\alpha_{\mathrm{sh}}\operatorname{Std}_{\mathrm{id}}(E_{\mathrm{sh}}(x))$ ${}+\alpha_{\mathrm{tax}}\operatorname{Std}_{\mathrm{id}}(E_{\mathrm{tax}}(x))$, where the $\alpha$ terms are score-level weights. CUBE-E further uses $S_{\mathrm{CUBE-E}}(x)=S_{\mathrm{CUBE}}(x)+\alpha_{\mathrm{geo}}\operatorname{Std}_{\mathrm{id}}(E_{\mathrm{geo}}(x))$. A sample is rejected as unknown when its corresponding score exceeds threshold $\tau_{score}$, otherwise it is assigned the top-scoring species under $\tilde z_c(x)$ for CUBE or $\tilde z_c^{\mathrm{ctx}}(x)$ for CUBE-E.

\section{Experiments}

\begin{figure*}[tp]
\centering
\includegraphics[width=\textwidth]{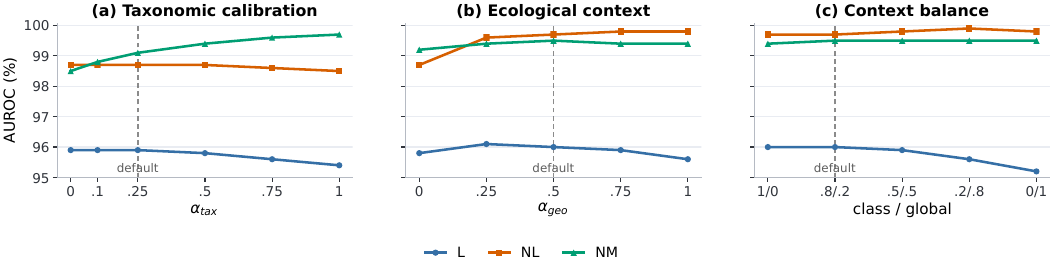}
\caption{Sensitivity of the three most influential scoring hyperparameters on C-America. Dashed vertical lines denote the default settings used in the main experiments.}
\label{fig:hyper_sensitivity}
\end{figure*}

\subsection{Experimental Setup}

We evaluate \method{} on the C-America split of Open-Insect~\cite{chen2026open}, using the \texttt{train\_id} split for closed-set training, the \texttt{train\_\allowbreak aux} split for auxiliary-aware methods, and the validation split for model selection and score standardization. We report closed-set accuracy (Acc.) on the \texttt{test\_id} split and OOD AUROC on local (L), non-local (NL), and non-moth (NM) unknown splits. 
We further report AUROC, AUPR, and FPR95 on the Barro Colorado Island (BCI) OOD split for cross-source evaluation, together with closed-set accuracy (Acc.) on the \texttt{test\_id} split.
All results for \method{} and \methode{} are averaged over five random seeds. All methods in the main comparison follow the same C-America protocol and use ConvNeXt-T with $224\times224$ inputs. \method{}, NovelBranch, and Extended are trained for 15 epochs using AdamW, ID/auxiliary batch sizes of 96/96, backbone/head learning rates of $3\times10^{-5}$/$2\times10^{-4}$, and cosine decay to $10^{-6}$. MixOE follows its original 10-epoch mixed-sample recipe with batch sizes of 32/32 and learning rate $10^{-3}$, while post-hoc methods share the same trained ConvNeXt-T closed-set classifier.

\begin{table*}[tp]
\centering
\setlength{\tabcolsep}{2.5pt}
\renewcommand{\arraystretch}{1.13}
\begin{minipage}[t]{0.485\textwidth}
\vspace{0pt}
\caption{Scoring weights on C-America (defaults in bold).}
\label{tab:score_weights}
\smallskip
\small
\begin{tabular*}{\linewidth}[t]{@{\extracolsep{\fill}}cccc@{}}
\toprule
\multicolumn{2}{c}{Boundary evidence} & \multicolumn{2}{c}{Support calibration} \\
\cmidrule(r){1-2}\cmidrule(l){3-4}
Weight & Acc. (\%) & Weight & Acc. (\%) \\
\midrule
0.00 & 94.9 & 0.00 & 94.6 \\
0.05 & 94.9 & 0.25 & 94.8 \\
\textbf{0.10} & \textbf{94.9} & 0.50 & 94.8 \\
0.15 & 94.9 & \textbf{0.75} & \textbf{94.7} \\
0.20 & 94.9 & 1.00 & 94.8 \\
\bottomrule
\end{tabular*}
\end{minipage}\hfill
\begin{minipage}[t]{0.485\textwidth}
\vspace{0pt}
\caption{Neighborhood sizes on C-America (defaults in bold).}
\label{tab:neighbor_sizes}
\smallskip
\small
\begin{tabular*}{\linewidth}[t]{@{\extracolsep{\fill}}cccc@{}}
\toprule
& \multicolumn{3}{c}{Closed-set accuracy (\%)} \\
\cmidrule(l){2-4}
Count & Candidate top-$M$ & Support top-$R$ & Boundary neighbors \\
\midrule
1  & -- & 94.9 & 94.9 \\[0.43em]
3  & 94.7 & \textbf{94.7} & 94.9 \\[0.43em]
5  & 94.7 & 94.7 & \textbf{94.9} \\[0.43em]
10 & \textbf{94.7} & -- & 94.9 \\
\bottomrule
\end{tabular*}
\end{minipage}
\end{table*}

\begin{figure*}[tp]
\centering
\includegraphics[width=\textwidth]{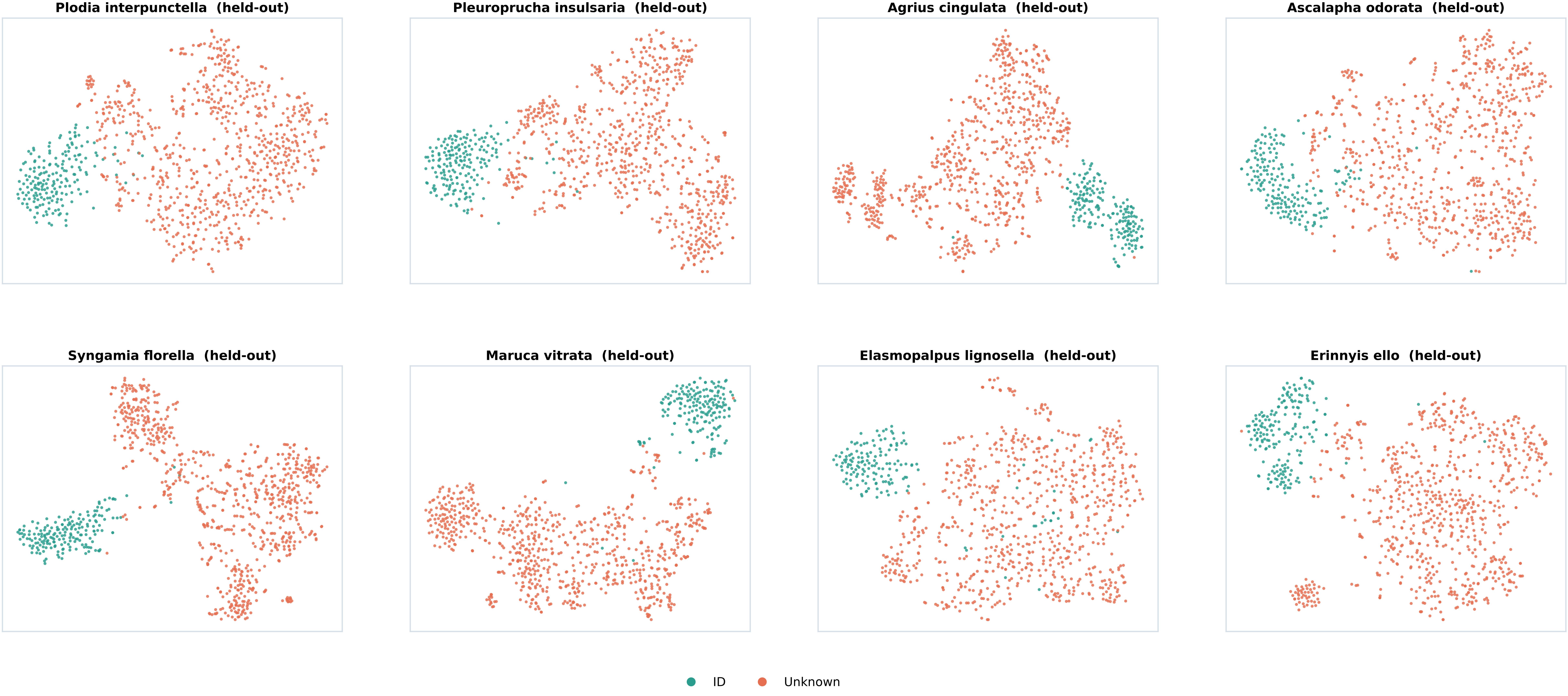}
\caption{Class-conditioned t-SNE visualization of \method{} evidence features. Each panel focuses on one held-out unknown species and its visually related ID neighborhood.}
\label{fig:tsne_class_conditioned}
\end{figure*}

\subsection{Main and Cross-Source Evaluation}

Table~\ref{tab:camerica_main} and Table~\ref{tab:bci} compare \method{} with ASCOOD~\cite{regmi2026image}, CADRef~\cite{Ling2025CADRefRO}, Mahalanobis++~\cite{mueller2025mahalanobispp}, OODD~\cite{yang2025oodd}, NECO~\cite{ammar2024neco}, NCI~\cite{liu2025detecting}, FDBD~\cite{liu2023fast}, MixOE~\cite{zhang2023mixture}, NovelBranch~\cite{chen2026open}, and Extended~\cite{chen2026open}. On C-America, \method{} improves Extended by 0.8, 0.4, and 0.8 AUROC scores on L, NL, and NM, respectively, with standard deviations of 0.4, 0.1, and 0.2, while \methode{} reaches 96.0, 99.7, and 99.5 AUROC with 95.2 accuracy. Across five matched seeds, \method{} significantly outperforms the strongest baseline on every metric in both evaluations. \methode{} is likewise significant except for BCI FPR95 (paired $t$-test, $p<0.05$). On the cross-source BCI split, \method{} obtains the best AUROC, AUPR, and FPR95 at $96.1\pm0.3$, $38.8\pm0.8$, and $16.1\pm1.2$. Overall, the core class-conditioned evidence generalizes across unknown sources, while ecological context helps when collection conditions are applicable.

\begin{figure*}[tp]
\centering
\includegraphics[width=0.85\textwidth]{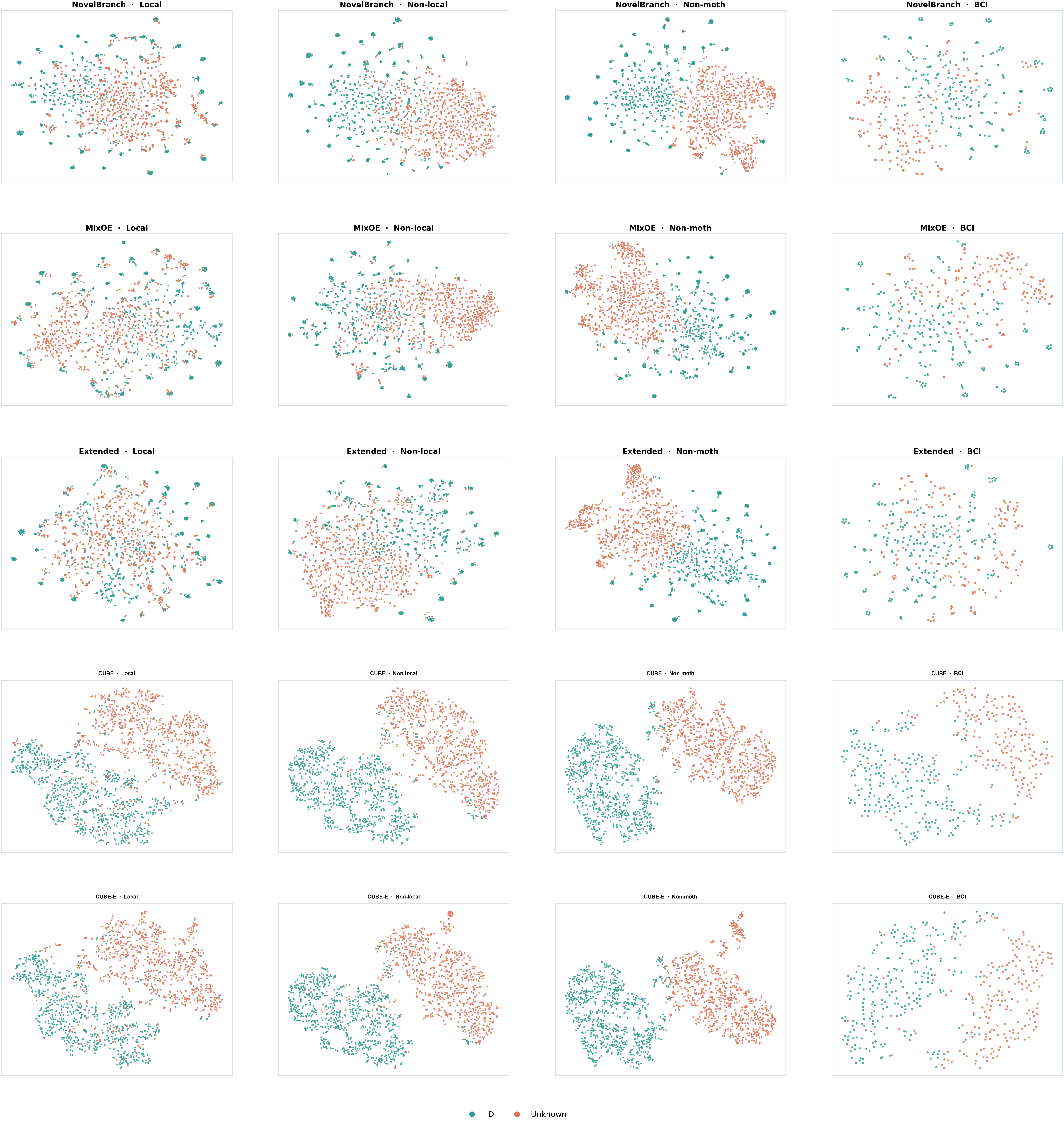}
\caption{t-SNE visualization across OOD sources. Rows correspond to NovelBranch, MixOE, Extended, \method{}, and \methode{}; columns correspond to local, non-local, non-moth, and BCI unknown splits.}
\label{fig:tsne_baseline_comparison}
\end{figure*}

\subsection{Ablation Study}

Table~\ref{tab:module_ablation} studies the contribution of the main components. Starting from the ID-only classifier, coarse auxiliary boundary learning ($\mathcal{L}_{\mathrm{mass}}$) raises L/NL/NM AUROC from 92.9/\allowbreak 94.7/96.7 to 94.9/97.9/98.1. Support-split boundary modeling ($\mathcal{L}_{\mathrm{exit}}$ and $\mathcal{L}_{\mathrm{br}}$) further improves L and NL to 95.6 and 98.5, while auxiliary species consistency ($\mathcal{L}_{\mathrm{asc}}$) increases closed-set accuracy from 94.1\% to 94.4\%. Preserving auxiliary prototypes as shadow evidence raises L/NL to 95.9/98.7, and taxonomic scoring raises NM from 98.5 to 99.2. Candidate calibration (w/o $\eta_{\mathrm{tax}}b_c(x)$) further improves closed-set recognition, yielding 94.9\% accuracy for the complete \method{}, together with 95.8/98.7/99.1 AUROC. Finally, \methode{} reaches 96.0/99.7/99.5 AUROC and 95.2\% accuracy by incorporating ecological consistency.

\subsection{Efficiency Analysis}

We measure efficiency under the online inference protocol used by the main experiments. All methods take a \(224\times224\) image and use ConvNeXt-T as the visual backbone; precomputed class text embeddings, auxiliary support banks, and ecological statistics are treated as cached evidence. Under this protocol, MSP-style, post-hoc, and training-regularized baselines share almost the same cost: 28.3M parameters, about 8.9G FLOPs, and 169.8--179.6 FPS on the RTX 3090. NovelBranch adds only a small auxiliary head, reaching 28.8M parameters and 178.6 FPS. Extended is the strongest auxiliary-branch baseline and is slightly heavier, with 31.9M parameters, 8.9G FLOPs, and 175.2 FPS.

With cached evidence, \method{} and \methode{} require 31.9M parameters and 8.9G FLOPs, matching existing baselines in complexity, and still achieve 77.2 and 80.7 FPS, respectively, even when BioCLIP features are recomputed online.

\subsection{Hyperparameter Analysis}

\noindent\textbf{Taxonomic scoring weight.} Figure~\ref{fig:hyper_sensitivity}(a) shows a clear trade-off: larger taxonomic evidence improves NM, while overly large weights slightly reduce L and NL. The default value 0.25 is therefore a conservative balance rather than a split-specific maximum.

\noindent\textbf{Ecological context weight.} Figure~\ref{fig:hyper_sensitivity}(b) shows that ecological context consistently improves NL and NM over the zero-context case. The gain saturates around 0.5, after which L mildly declines, matching its role as a useful but non-dominant location prior.

\noindent\textbf{Ecological class/global balance.} Figure~\ref{fig:hyper_sensitivity}(c) varies the mixture between class-specific and global geography. More global weight slightly improves NL but hurts L, while the default 0.8/0.2 balance keeps local performance high and preserves the NL/NM gains.

\noindent\textbf{Boundary evidence weight.} Table~\ref{tab:score_weights} shows that varying this weight from 0 to 0.2 keeps closed-set accuracy at 94.9\%. Mean OOD AUROC is 97.9\% across all settings reported in Tables~\ref{tab:score_weights} and~\ref{tab:neighbor_sizes}. Thus, the representation already encodes most boundary evidence, and this weight only provides a mild correction.

\noindent\textbf{Support calibration weight.} This weight affects closed-set candidate calibration but leaves the OOD scorer fixed. The accuracy range is narrow, from 94.6 to 94.8, and the OOD average is unchanged, supporting its role as a local classification correction.

\noindent\textbf{Candidate class count.} The candidate top-$M$ controls how many visually plausible species are considered during taxonomic candidate calibration, as shown in Table~\ref{tab:neighbor_sizes}. Using 3, 5, or 10 candidates gives the same OOD average and accuracy, so the correction remains local.

\noindent\textbf{Support aggregation count.} The support top-$R$ controls how many anchors are averaged for each candidate. The results are stable from 1 to 5 anchors, with only a 0.2-point accuracy range, suggesting that the support bank captures class neighborhoods robustly.

\noindent\textbf{Boundary neighbor count.} Sweeping the number of known species that receive local boundary supervision from 1 to 10 produces identical aggregate metrics in Table~\ref{tab:neighbor_sizes}. The learned boundary is therefore not brittle to the exact neighborhood size.

\subsection{Feature Visualization}

We further visualize the learned evidence space with t-SNE. Teal points denote ID images and coral points denote unknown images. These figures are not used for model selection; they are intended to qualitatively inspect whether the learned evidence separates visually similar unknown samples from known species.

Figure~\ref{fig:tsne_class_conditioned} shows that unknown samples near specific held-out species form regions separated from the corresponding ID neighborhoods in most panels. This behavior is consistent with the class-conditioned design of CUBE: the model does not only push all unknown samples away globally, but also learns local evidence around individual known-species neighborhoods.
Figure~\ref{fig:tsne_baseline_comparison} further compares strong auxiliary baselines with our evidence features under the same four OOD sources. The baseline rows often contain mixed ID/OOD regions, especially on local and non-local unknowns. In contrast, \method{} produces clearer two-region structures, and \methode{} further improves separation on the non-local split.

\section{Conclusion}

In this paper, we presented CUBE, a class-conditioned framework for open-set insect species recognition. Instead of modeling unknownness globally, CUBE learns local boundaries from auxiliary species, preserves auxiliary prototypes as inference-time shadow references, and uses taxonomic evidence to assess biological plausibility. CUBE-E further incorporates optional ecological context. Extensive experiments on the C-America and cross-source BCI splits demonstrate that our framework can improve unknown rejection while maintaining strong closed-set recognition, supporting class-conditioned biological evidence as an effective approach for fine-grained open-set recognition.

\section{Acknowledgements}
This work was supported in part by the National Natural Science Foundation of China (NSFC) under Grant No. 62206314, Science and Technology Projects in Guangzhou under Grant No. 2024A04J4388.

\bibliographystyle{IEEEbib}
\bibliography{strings,refs}

\end{document}